# TMR-GGNN: Credit Card Fraud Detection based on Time-Aware Multi-Relational Guided Graph Neural Network

1st Rohit Tewari
*Unysis*
Fairfax, USA
rohittewari.fintech@gmail.com

2nd Shubhankar Shilpi
*Truist Banks*
Atlanta, USA
shubhankarshilpi.tech@gmail.com

3rd Navin Chhibber
*Infinity Tech Group Technical Product Owner*
Sunnyvale, USA
naveenchibber.research@gmail.com

4th Devendra Singh Parmar
*Discover Financial Services*
Chicago, USA
davesingh081@gmail.com

5th Sunil Khemka
*Persistent Systems*
Chicago, USA
sunilkhemka.tech@gmail.com

6th Piyush Ranjan
*IEEE Vice Chair AeroSpace Chapter*
Edison, USA
piyush.ranjan@ieee.org

***Abstract*—In recent years, credit card fraud detection has faced significant challenges due to highly imbalanced data, evolving fraud patterns, and complex relational structures among transaction entities. To address these issues, this research proposes a novel framework called Time-aware Multi-Relational Guided Graph Neural Network (TMR-GGNN). Particularly, the proposed TMR-GGNN extends the encoder–decoder Graph Neural Network GNN architecture by modeling heterogeneous interactions across customers, merchants, devices, and IPs over temporal windows. Subsequently, the proposed TMR-GGNN approach constructs a dynamic, multi-relational graph and incorporates a time-aware relational attention mechanism within the encoder to adaptively weigh the transaction relevance based on temporal proximity and semantic context. Consequently, the decoder employs a contrastive learning module to distinguish between real and synthesized transaction patterns, while improving the model's generalization of rare fraud cases. Additionally, to effectively manage severe class imbalances and emphasize discriminative learning, a composite loss function combining Information Noise-Contrastive Estimation (InfoNCE)-based contrastive loss with Focal Loss is introduced. This integration assists in improving fraud identification while mitigating false negatives.**

## I. INTRODUCTION

The rapid digitization of financial services integrated with an exponential growth of online transaction activity has significantly increased the risk of payment systems particularly in terms of credit card fraud [1]. This threat continues to escalate in an increasingly dynamic global financial landscape, exceeding billions of losses with emerging attack patterns which is becoming more complicated and difficult to detect [2]. Traditional rule-based systems are an important foundation but they are not sufficient to tackle the complications of modern fraud tactics. One of the primary issues in the field of credit card fraud detection is the class imbalance, where the number of fraudulent transactions is extremely small compared to the total number of activities, which makes it more difficult for regular Machine Learning (ML) systems to mine possible patterns [3]. Additionally, the massive volume of transaction data and varying fraud behaviors make the manual observation of fraudulent patterns unreasonable. To resolve these issues, recent research has focused on Deep Learning (DL) approaches, which have resulted in better harnessing of nonlinear and temporal fraudulent behaviors [4].

Furthermore, conventional models have limitations in detecting unknown attack behaviors, thus requiring AI approaches with greater sophistication for fraud detection [5]. Oversampling technique may also introduce noise, whereas unsupervised methods run the risk of high false positive rates, all of which restrict the reliability and generalization of detection systems [6]. The increasing size of credit card fraud and issues of subtle attacks identification illustrates a significant lack of fraud detection. [7]. Moreover, the misclassification, data imbalance, and limitations of existing models necessitate sophisticated methods to enhance the detection accuracy and reliability of fraud-detection systems [8]. Recent research has investigated multiple fraud-detection models, such as Support Vector Machines (SVM) and neural networks, and methods such as ensembles that deal with the imbalanced data [9]. Particularly, graph-based models such as Graph Neural Networks (GNNs), Variational Graph Autoencoders (VGAEs), and new techniques such as GraphSMOTE and ADA-GAD can improve the identification of anomalous actions in complex transactional networks [10]. The key contributions of this research are:

- Heterogeneous graph modeling for fraud detection, which is a multi-relational graph constructed to capture the interactions among customers, merchants, and devices that allowed relational and structural modeling of transaction behavior.
- A time-aware relational attention mechanism integrated time decay into the graph transformer, thereby enabling the model to emphasize temporally relevant interactions during message passing.
- A guided contrastive learning decoder utilized contrastive learning to improve the embedding separability and the ability of the model to distinguish between fraudulent and legitimate transaction patterns.
- The Composite Loss Function is employed to resolve extreme class imbalance, which combined focal loss and Information Noise-Contrastive Estimation (InfoNCE)-based contrastive loss to improve the accuracy and representation quality.

The organization of this research is structured as follows: Section 2 describes the literature review, Section 3

demonstrates the proposed methodology, Section 4 illustrates the experimental results, and Section 5 concludes the paper.

## II. Literature Review

Asma Cherif et al. presented an encoder–decoder GNN for credit card fraud detection [11]. This approach suggested utilizing GNNs to model the complex relationships between customers and merchants. The objective was to handle class imbalance, evolving fraud tactics, and adversarial manipulation. Thus, the proposed model improved the precision, recall, and F1 score by modeling structural graph features. However, it was ineffective due to its dependency on high-quality graph data and potential scalability challenges for real-time deployment.

Georgios Charizanos et al. recommended an online fuzzy fraud detection framework using fuzzy logistic regression [12]. This framework addressed challenges such as class imbalance, non-stationarity, and a complete separation of credit card transaction data. The objective was to achieve a high accuracy and real-time adaptability. Their method achieved higher specificity and sensitivity than the other ML techniques. Nonetheless, this framework relied on a narrow performance metric set and lacked the evaluation of diverse dataset.

Yuxuan Tang and Zhanjun Liu suggested a Structured Data Transformer (SDT) integrated with federated learning for fraud detection [13]. Their objective was to overcome the limitations of traditional models in handling serialization, feature engineering complexity, and data privacy. This model utilized attention mechanisms and distributed training to enhance bank detection. Although the model achieved excellent AUC-ROC and AUC-PR scores, it was tested only in emulated environments, thereby limiting its validation in actual production systems.

Zhichao Xie and Xuan Huang presented a method using Mahala Nobis Distance SMOTE-ENN hybrid sampling with Random Forest for fraud detection [14]. They recommended addressing class imbalances by utilizing refined sampling and enhanced model robustness. The objective was to retain the minority class characteristics while minimizing overfitting. This method achieved superior performance metrics compared to other ML methods. However, the dependency on predefined sampling parameters resulted in inefficiency when generalized to other data types.

Emmanuel Ileberi and Yanxia Sun recommended a hybrid DL ensemble model that integrated CNN, LSTM and Transformer architectures, with XGBoost as a meta-learner [15]. This hybrid model aimed to resolve the issues of imbalanced dataset and evolving fraudulent strategies. The ensemble model captured the spatial, temporal, and attention-based dependencies, thereby achieving high specificity and AUC-ROC values. However, the use of a single older dataset limited generalizability, and was inefficient while getting adapted to new fraud behaviors in diverse demographics.

## III. Methodology

The proposed Time-aware Multi-Relational Guided Graph Neural Network (TMR-GGNN) framework constructs a dynamic and a multi-relational representation based on customer credit card transactions to model several entities, as demonstrated in Fig. 1. Specifically, the model employs a time-aware relational graph attention encoder which captures both semantic and temporal dependencies by adjusting the attention weights according to the edge types and time decay functions. For an effective fraud detection in terms of patterns, a guided decoder with a contrastive prediction head is utilized to distinguish between actual transactions and synthetically generated negatives through structure-preserving random walks. The model is trained by utilizing an integrated loss function, namely InfoNCE-based contrastive loss and Focal Loss, which considers structure-based learning within the relation graph and addresses parametric class imbalance. Hence, this end-to-end architecture provides fraud detection across multiple transactions in a single complex interaction representation that is interpretable and cost-sensitive.

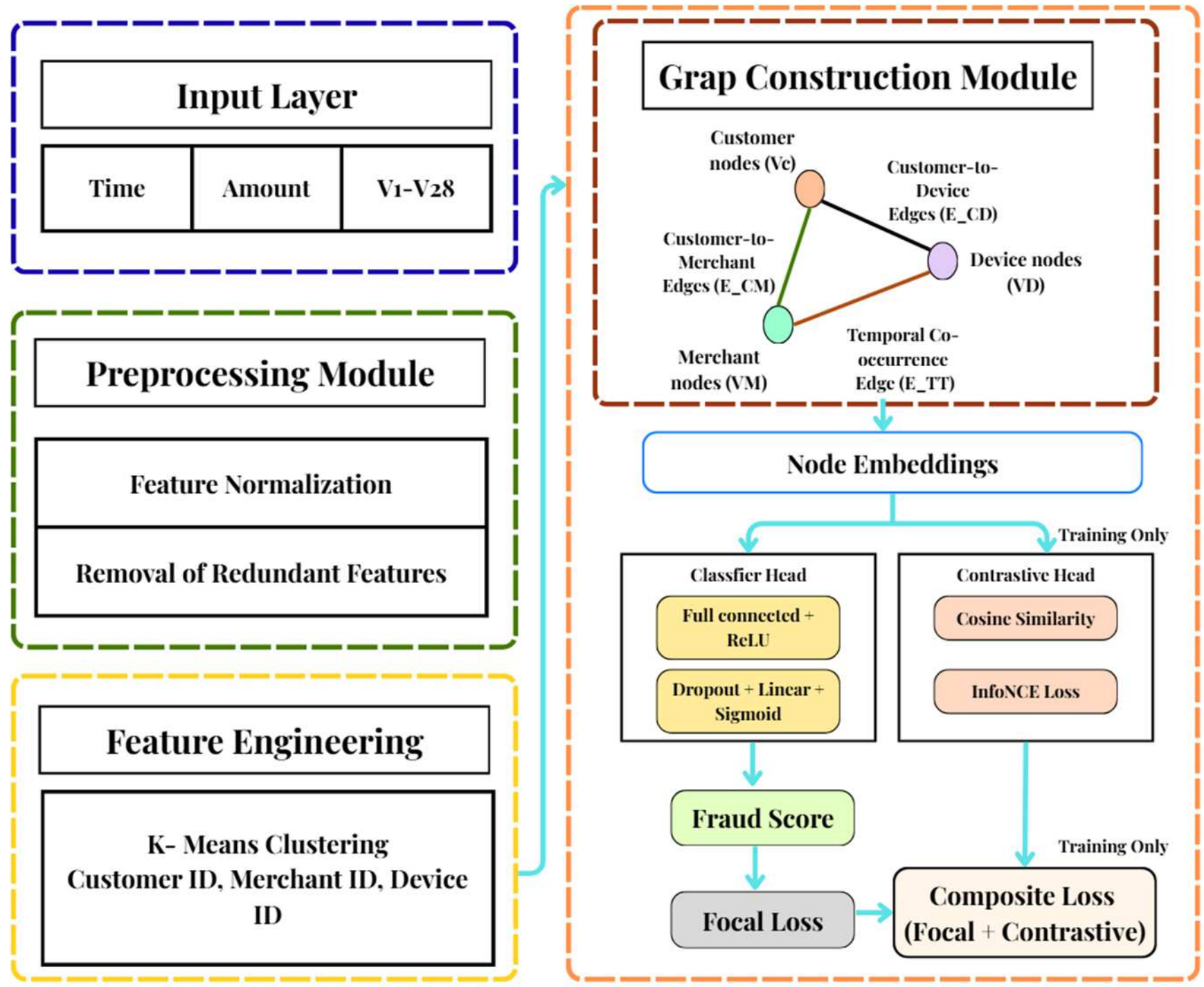

Fig. 1. Overall architecture of the proposed TMR-GGNN for credit card fraud detection

## A. Data Acquisition

This research employs the well-known European credit card transactions dataset which has anonymized credit card transactions from European card holders in September 2013 [12]. Specifically, the dataset contains 284,807 transactions, with only 492 (0.172%) having a fraud label representing a typical extremely imbalanced distribution of classes in financial fraud detection. Subsequently, the transaction instances are made up of 30 features containing numerical data with features V1 to V28 obtained from a Principal Component Analysis (PCA) transformation for privacy reasons, with the additional features of time, amount, and class. Hence, the dataset illustrates heterogeneous relational patterns, including different entities, such as cardholders and merchants, and the context of transacting and enriching these patterns by employing metadata, such as pseudo-identifiers for customers, merchants, temporal buckets, and derived device/IP cluster identifiers. Henceforth, these enrichments allow the development of a temporal, multi-relational graph that is useful for graph-based learning models. Thus, this dataset reflects the structural and behavioral diversity of financial transactions and provides a foundation for constructing a heterogeneous graph.

## B. Data preprocessing

In this stage, a sequence of essential preprocessing steps is employed to standardize and clean the dataset. These steps ensure numerical stability, eliminate redundant information, and facilitate accurate downstream clustering and modeling. The major preprocessing steps include feature normalization and removal of non-informative attributes, as detailed below.

*1) Feature Normalization:* The normalization technique is employed to rescale continuous numeric features, such that each variables contribute proportionally during clustering. Subsequently, normalization ensures that each feature, particularly those with different ranges, are placed on a comparable scale. The normalized is employed using two techniques, as represented in Eqs. (1) and (2).

$$x_i^{norm} = \frac{x_i - \mu}{\sigma} \quad (1)$$

Where $x_i$ is the original feature value defined as $x_i$, $\mu$ is the mean of the features represented as $\mu$ and $\sigma$ is the standard deviation.

$$x_i^{norm} = \frac{x_i - x_{min}}{x_{max} - x_{min}} \quad (2)$$

Where $x_{min}$, $x_{max}$ represent the minimum and maximum feature values, respectively. Specifically, the Z-score is preferred for normally distributed variables, whereas Min-Max is ideal when data must be bounded within a fixed interval of [0,1].

*2) Removal of Redundant Features:* This technique is incorporated to eliminate columns that do not provide significant information for pattern recognition or relationship modeling. Specifically, attributes, such as transaction ID, are considered as unique row identifiers with no statistical relationship to the target label or other features. Similarly, features with near-zero variance provide no discriminatory power during learning, which also increases noise and computational costs. Hence, feature $f_j$ is considered non-informative and is removed if. $Var(f_j) \approx 0 \ or \ | \ Unique(fj) \ | \approx N$

Here, the variance of feature $f_j$ is denoted as $Var(f_j)$, the number of unique values is represented as $Unique(fj)$ and $N$ signifies the total number of data instances. Thus, removing such columns simplifies the feature space and prevents overfitting during the model training.

## C. Feature Engineering

In addition to the standard normalization and cleaning steps, the number of domain-based features which helped with the construction of the graph were designed. Because the original dataset is anonymized and does not contain a unique identifier, this research design comprised fields such as Customer ID and Merchant ID that are created by clustering the anonymized PCA variables which indicate the transaction patterns. Further, Device/IP Cluster is created by clustering transactions around temporal-spatial similarities and adding time bucket features to discretize the time variable into hourly blocks.

K-means clustering is applied to different subsets of the original PCA-transformed features. Given a dataset $X = \{x1, x2, \ldots, xn\} \subset \mathbb{R}^d$, the goal of K-Means is to partition the data into $k$ disjoint clusters which are $C1, C2, \ldots, Ck$ by minimizing the within-cluster sum of squared distances (WCSS) as demonstrated in the Eq. (3):

$$arg \underset{C}{min} \sum_{j=1}^{k} \sum_{x_i \in C_j} \left\| x_i - \mu_j \right\|^2 \quad (3)$$

Where the centroid of cluster $C_j$ is defined as $\mu j = \frac{1}{|Cj|} \sum_{xi \in C_j} x_i$ , the squared Euclidean distance between point $x_i$ and its cluster centroid is denoted as $\left\| x_i - \mu_j \right\|^2$ and $C = \{C_1, \ldots, C_k\}$ represents the set of all clusters.

Furthermore, each data point $x_i$ is assigned a cluster label, as shown in Eq. (4).

$$Cluster\ (x_i) = arg \underset{j}{min} \left\| x_i - \mu_j \right\|^2 \quad (4)$$

Furthermore, the Silhouette Coefficient, which is a standard metric for assessing clustering quality was employed to determine the optimal number of clusters $k$ for generating synthetic identifiers. For each data point, the silhouette score $s(i)$ measures how well it fits within its assigned cluster compared to neighboring clusters using Eq. (5):

$$s(i) = \frac{b(i) - a(i)}{max\{a(i), b(i)\}} \quad (5)$$

Where $a(i)$ is the mean intra-cluster distance (cohesion) and $b(i)$ is the mean nearest-cluster distance (separation). The coefficient ranges from $-1$ to 1 with higher values indicating better-defined clusters. Hence, the $b$ is evaluated over all points for a range of $k$ values, the $k$ is selected to provide the maximum silhouette score and facilitate the balancing of compactness and separation of clusters. Hence, K-means is applied distinctly to feature subsets to generate synthetic entity identifiers, as shown in Table 1.

From Table 1, it is inferred that the clusters of transactions representing customers which are created using the first 10 principal components $(V1 - V10)$, whereas merchant-like behavior is assigned through the clustering of either $V11 - V20$.

For approximate shared device/IP contexts, K-means clustering is performed on a low-dimensional projection of all $V1 - V28$ components, as well as the normalized time and amount. Subsequently, each cluster label is assigned as a categorical identifier (Customer ID, Merchant ID, Device/IP Cluster) that assists in constructing the nodes in the relational graph. Hence, this consistent and scalable clustering approach across sections allows the creation of entity-level interactions in the model. A brief description of these representations created with the original dataset features is summarized in Table 2.

TABLE I. FEATURE VECTOR FORMULATIONS AND RESULTING SYNTHETIC LABELS FOR ENTITY-LEVEL IDENTIFIERS

| Entity Type | Input Feature Vector $x_i$ | Resulting Label |
|---|---|---|
| Customer ID | $x_i = [V1, V2, \ldots, V10] \in \mathbb{R}^{10}$ | CustomerID ($x_i$) |
| Merchant ID | $x_i = [V11, \ldots, V20] \in \mathbb{R}^{10}$ | MerchantID ($x_i$) |
| Device/IP Cluster | $x_i = [PCA_{1-2}(V1 - V28), time, Amount] \in \mathbb{R}^3$ | DeviceID($x_i$) |

TABLE II. SUMMARY OF CLUSTERING-BASED FEATURE ENGINEERING FOR SYNTHETIC ENTITY IDENTIFIERS USED IN GRAPH CONSTRUCTION

| Feature Name | Type | Description |
|---|---|---|
| Time | Numerical | Time elapsed in seconds since the first recorded transaction in the dataset |
| V1 to V28 | Numerical | Principal components obtained via PCA to anonymize sensitive customer features |
| Amount | Numerical | Monetary value of the transaction in Euros |
| Class | Categorical | Target label indicating transaction type: 1 for fraud, 0 for legitimate |
| Customer ID | Categorical | Unique pseudo-identifier for the cardholder |
| Merchant ID | Categorical | Unique pseudo-identifier for the merchant involved in the transaction |
| Transaction ID | Categorical | Unique identifier for each transaction used for indexing |
| Device/IP Cluster* | Categorical | Grouped feature representing shared IP or device types |
| Time Bucket | Categorical | Discretized time interval grouping for temporal modeling |

Hence, these engineered features assist in creating a heterogeneous graph that includes multiple node types with edges that are based on time, allowing the model to learn patterns related to certain behaviors while also learning the relational dependencies of fraudulent transactions.

### *D. Graph Construction*

To effectively capture the complex relational structure that is essential for credit card transactions, the dataset is designed as a heterogeneous and multi-relational graph. This operation enables the GNN to learn high-level fraud patterns through entity interactions, behavioral dependencies, and temporal dynamics.

*1) Node Definition:* There are three types of nodes to define the key entities in the transaction patterns, which are described as follows:

- Unique customers derived from synthetic Customer ID clusters are represented as Customer nodes ($\mathcal{V}_C$).
- Individual merchant derived through Merchant ID clustering is defined as ($\mathcal{V}_M$).
- Shared access, such as devices or IP addresses, is determined by device/IP nodes ($\mathcal{V}_D$).

The full node set is defined as expressed in the Eq. (6):

$$\mathcal{V} = \mathcal{V}_C \cup \mathcal{V}_M \cup \mathcal{V}_D \quad (6)$$

*2) Edge Definitions:* The edges determine the relationship between entities based on various transaction patterns, as defined below.

- Customer-to-Merchant Edges ($\mathcal{E}_{CM}$): Each transaction establishes an edge from the customer node to the associated merchant node.
- Customer-to-Device Edges ($\mathcal{E}_{CD}$ ): Edges are created from customers to their associated device/IP nodes.
- Temporal Co-occurrence Edges ($\mathcal{E}_{TT}$): An undirected edge is added between any two transactions that occur within a time window (e.g., 1 h) that share a merchant.

Hence, each edge $e_{ij} \in \mathcal{E}$, is represented as triplet through Eq. (7):

$$e_{ij} = (v_i, v_j, r_{ij}) \quad (7)$$

Where, $v_i, v_j \in \mathcal{V}$ signifies the source and target nodes and $r_{ij} \in \mathcal{R}$ symbolizes the edge types.

Thus, the full edge set is defined as expressed in Eq. (8):

$$\mathcal{E} = \mathcal{E}_{CM} \cup \mathcal{E}_{CD} \cup \mathcal{E}_{TT} \quad (8)$$

*3) Edge Features and Temporal Encoding:* In particular, each edge comprises contextual features such as transaction amount, time difference between adjacent events, and edge-type encoding. Thus, for temporal modeling, each edge is assigned a time bucket based on discretized time intervals, which allows the model to learn the time-aware attention weights in the GNN encoder.

Hence, the final graph constructed is defined using Eq. (9):

$$\mathcal{G} = (\mathcal{V}, \mathcal{E}, \mathcal{R}, X) \quad (9)$$

Where the set of nodes is denoted as $\mathcal{V}$, the set of edges is determined as $\mathcal{E}$, the set of edge relations is denoted as $\mathcal{R}$ and $X$ denotes the node feature matrix comprising transaction-level embeddings. Hence, building a heterogeneous graph allows the model to capture more complicated relations and temporal patterns that have been lost in a tabular form. The graph includes three different nodes for a customer, merchant, and device with an edge connecting them and models the behavioral dependencies and combined patterns vital for fraud detection. Henceforth, the addition of temporal data and edge type further enhances the model's learning by allowing it to learn fraud behavior dynamically, model correlations between entities, and provide more accurate and context-aware fraud detection.

*E. Model Architecture*

For an effective modeling of the heterogeneous and temporal relationships in the credit card fraud dataset, a modified Graph Transformer Network (GTN) architecture is utilized, which includes a time-aware relational encoder and guided contrastive decoder. Specifically, this structure enables the model to learn discriminative representations by aggregating node interactions across multiple relation types, while also capturing long-range dependencies and temporal behaviors included in credit card fraud patterns. This section outlines the baseline model and presents key innovations introduced to enhance fraud detection performance under diverse conditions, such as extreme class imbalance, relational uncertainty, and temporal variation.

GTN is a deep learning architecture designed to operate on heterogeneous graphs with multiple edge types. It generalizes an attention-based learning to graph-structured data by learning soft edge-type weights and performing automatic relation-path learning. The encoder aggregates information across multi-relational edges using attention scores, thereby enabling a rich semantic representation. The decoder processes the resulting embeddings for downstream prediction tasks. Subsequently, transformer-like attention is utilized for relational graphs, also allowing the dynamic learning of important relation paths. Hence, GTN supports multi-hop reasoning over heterogeneous edge types, making it well suited for our transaction graph with customer–merchant–device relations.

In the encoder, for a node $v_i \in \mathcal{V}$, the model computes attention-based aggregation from its multi-relational neighborhood $v_i \in \mathcal{N}_i^{(r)}$ for each edge type $e_{ij} = (v_i, v_j, r_{ij}) \in \mathcal{E}$. The relation-specific attention coefficient was evaluated using Eq. (10).

$$\alpha_{ij}^{(r)} = \frac{exp\left(LeakyReLU\left(a_r^T(W_r h_i \| W_r h_j)\right)\right)}{\sum_{k \epsilon \mathcal{N}_i^{(r)}} exp\left(LeakyReLU\left(a_r^T(W_r h_i \| W_r h_k)\right)\right)} \quad (10)$$

Where the embedding of node $v_i$ is defined as $h_i \epsilon \mathbb{R}^d$, the relation specific transformation matrix is denoted as $W_r$, the learnable attention vector for relation type $r \epsilon \mathcal{R}$ is determined as $a_r$ and $\|$ signifies the concatenation. The updated model embedding is assessed as demonstrated in the Eq. (11):

$$h_i' = \sigma(\sum_{r \epsilon \mathcal{R}} \sum_{v_i \epsilon \mathcal{N}_i^{(r)}} \alpha_{ij}^{(r)} \cdot W_r h_j) \quad (11)$$

Thus, the attention mechanism learns to distinguish between the types, as in credit card fraud graphs, and the interaction varies in importance based on the type. Furthermore, the decoder utilizes the final embedding $h_i'$ to estimate fraud likelihood, as shown in Eq. (12):

$$\hat{y}_i = Sigmoid(W_\circ \cdot h_i' + b) \quad (12)$$

Where, the predicted fraud probability for node $v_i$ is defined as $y^i \in [0,1]$, sigmoid activation function $\sigma(x) = \frac{1}{1+e^{-x}}$ is mentioned as $Sigmoid(\cdot)$, final embedding vector of node $v_i$ is denoted as $h_i' \epsilon \mathbb{R}^d$, the output weight matrix that maps embedding to a scalar logic is represented as $W_\circ \epsilon \mathbb{R}^{1 \times d}$ and $b \in \mathbb{R}$ signifies the bias term applied after linear transformation. Thus, this maps the node embedding to a scalar fraud score and then applies the sigmoid function to convert it into a probability.

*1) Time-Aware Relational Graph Attention Encoder:* To integrate the transaction timing, the attention mechanism is modified by incorporating a temporal decay over the time difference $\Delta t_{ij}$ associated with edge $e_{ij}$ as expressed in Eq. (13).

$$\alpha_{ij}^{(r)} = \frac{exp\left((W_q h_i)^T (W_k h_j) \cdot exp(-\gamma \cdot \Delta t_{ij})\right)}{\sum_{k \epsilon \mathcal{N}_i^{(r)}} exp\left((W_q h_i)^T (W_k h_k) \cdot exp(-\gamma \cdot \Delta t_{ik})\right)} \quad (13)$$

Where the learnable temporal decay coefficients are denoted as $\gamma$, $W_q$ ,$W_k$ denote the projection matrices for the query and key, respectively. Hence, this mechanism prioritizes recent interactions and down-weights outdated interactions by correlating fraudulent behavior and assists the model in dynamically emphasizing relevant edges.

*2) Guided Contrastive Decoder (InfoNCE):* Further, a contrastive loss component is presented to assist in embedding separability. Specifically, positive pairs $(v_i, v_j)$ represent structurally or temporally similar nodes and negatives $(v_i, v_k)$ are sampled randomly or from different clusters. The contrastive loss is formulated as mentioned in Eq. (14):

$$\mathcal{L}_{InfoNCE} = -log \frac{exp(sim(h_i, h_j)/\tau)}{\sum_{v_k \epsilon \mathcal{N}_i^-} exp(sim(h_i, h_k)/\tau)} \quad (14)$$

Where the contrastive loss for a given positive pair $(v_i, v_j)$ is defined as $\mathcal{L}_{contrastive}$ , embedding of negative sample nodes $v_k$ , set of negative samples for anchor node $v_i$ is demonstrated as $\mathcal{N}_i^-$, the similarity function typically cosine similarity $sim(h_a, h_b) = \frac{h_a' \cdot h_b}{\|h_a\| h_b\|}$ is defined as $sim\ (.,.)$, the temperature scaling parameter typically $\tau \epsilon$ [0.05,0.5],helps to control separation sharpness and $exp(\cdot)$ signifies the exponential function which assists to convert similarities to unnormalized score. Hence, this loss ensures the anchor node $v_i$ to be adjacent in embedding space to its positive sample $v_j$ and distant from all negatives $v_k \epsilon \mathcal{N}_i^-$ . Additionally, the temperature $\tau$ controls how sharply the model focuses on hard negatives. Because fraud is often established in dense substructures, contrastive learning is employed because it helps the model in differentiating such patterns beyond binary labels and enforces the clustering of fraud-relevant behavior in the embedding space.

*3) Composite Loss Function:* Furthermore, a combined loss to handle both class imbalance and embedding quality is designed, as formulated in Eq. (15):

$$\mathcal{L}_{total} = \lambda_1 \cdot \mathcal{L}_{InfoNCE} + \lambda_2 \cdot \mathcal{L}_{focal} \quad (15)$$

Where the focal loss ($\mathcal{L}_{focal}$) computed for a single training instance modifies the standard cross-entropy to focus on difficult (misclassified) samples, as defined in Eq. (16).

$$\mathcal{L}_{focal} = -\alpha_t (1 - p_t)^\gamma log(p_t) \quad (16)$$

Where, Class balancing factor $\alpha_t$ is used to address class imbalance by giving more weight to the minority class that typically set higher for fraud class, the predicted probability of the true class $y_i$ for instance $i$ obtained from the sigmoid output in binary classification is represented as $p_t = \hat{y}_i \epsilon (0,1)$, focusing parameter that adjusts the rate at which easy examples are down-weighted and symbolized as $\gamma \geq 0$ and $log(p_t)$ signifies the log-likelihood of the true class. Hence, the proposed model detects fraud by learning context-aware embeddings from a heterogeneous transaction graph using time-aware attention and contrastive guidance. Finally, predictions are made using a decoder optimized with a composite loss to handle both class imbalance and structural variability.

## IV. Experimental Results

All experiments utilizing the system are equipped with an NVIDIA RTX 3090 GPU (24GB VRAM), 128GB RAM, and AMD Ryzen 9 5950X CPU. The implementation is executed using Python 3.10, with the core modeling performed in PyTorch 2.0 and PyTorch Geometric (PyG), with which graph-based learning capabilities are employed. Further, the additional libraries utilized are NumPy and scikit-learn to evaluate the research models. The proposed TMR-GGNN is evaluated in terms of accuracy, precision, recall f1-score, and AUC-ROC, as formulated in Eqs. (17) – (20):

$$Accuracy = \frac{TP+TN}{TP+FP+FN+TN} \quad (17)$$

$$Precision = \frac{TP}{TP+FP} \quad (18)$$

$$Recall = \frac{TP}{TP+FN} \quad (19)$$

$$F1 - Score = \frac{2.Precision.Recall}{Precision+Recall} \quad (20)$$

Where $TP$ denotes True Positives, $FP$ represents False Positives, $FN$ signifies False Negatives, $TN$ denotes True Negatives.

### A. Performance Analysis

To validate the effectiveness of the proposed TMR-GGNN, its performance was analyzed against several state-of-the-art baseline models such as LSTM, CNN, and GCN, as illustrated in Table 3.

TABLE III. Performance Analysis Of Proposed TMR-GGNN With State-Of-The-Art Baseline Models

| Model | Accuracy | Precision | Recall | F1-Score | AUC-ROC |
|---|---|---|---|---|---|
| LSTM | 0.963 | 0.785 | 0.881 | 0.830 | 0.901 |
| CNN | 0.961 | 0.760 | 0.850 | 0.802 | 0.894 |
| Autoencoder | 0.958 | 0.745 | 0.832 | 0.786 | 0.880 |
| GCN | 0.965 | 0.800 | 0.865 | 0.831 | 0.910 |
| Proposed TMR-GGNN | 0.999 | 0.975 | 0.955 | 0.945 | 0.996 |

As shown in Table 3, the proposed TMR-GGNN outperforms all baseline models across all metrics, particularly in terms of recall (0.955) and F1-Score (0.945), which are essential for fraud detection. Hence, the TMR-GGNN maintains high precision and recall, ensuring better fraud coverage with minimal false alarms. This signifies the effectiveness of incorporating temporal attention and contrastive supervision into graph-based fraud detection.

### B. Comparative Analysis

To assess the effectiveness of the proposed TMR-GGNN model, a comparative analysis is performed with existing models, such as the encoder–Decoder GNN [11] and the SDT framework [12]. These models represent strong state-of-the-art baselines for graph- and transformer-based fraud detection, as presented in Table 4.

The results in Table 4 show that the TMR-GGNN Model clearly outperformed all evaluated metrics. Compared to the Encoder Decoder GNN [11], TMR-GGNN showed a significant improvement in recall (0.920–0.955) and F1-Score (0.860–0.945), which indicates that TMR-GGNN is better at detecting fraudulent transactions with fewer false negatives. The SDT [12] model shows high overall accuracy (0.997), but has a recall of lower recall (0.734), which means the SDT model demonstrated strong predictive ability overall, but missed fraud cases. This indicates that SDT has a tradeoff between overfitting the majority class and missing fraud in the predictions. TMR-GGNN achieved balanced overfitting based on the predictive performance of both high precision (0.975) and recall; therefore, TMR-GGNN is a potentially better solution for robust fraud detection.

TABLE IV. Performance Analysis Of Proposed TMR-GGNN With Existing Models

| Models | Accuracy | Precision | Recall | F1-Score | AUC-ROC |
|---|---|---|---|---|---|
| Encoder Decoder GNN [11] | 0.970 | 0.820 | 0.920 | 0.860 | 0.920 |
| SDT [12] | 0.997 | 0.961 | 0.734 | 0.754 | 0.994 |
| Proposed TMR-GGNN | 0.999 | 0.975 | 0.955 | 0.945 | 0.996 |

### C. Discussion

The Proposed TMR-GGNN architecture provides significant advantages over the existing state-of-the-art models for credit card fraud detection applications. Conventional DL models such as those LSTM, CNN and Autoencoders depend on flat feature representations, by which these models struggle to capture inter-entity relationships. Additionally, the TMR-GGNN utilizes a heterogeneous graph representation that models the interactions of complexity between customers, merchants, and devices. However, TMR-GGNN employs a time-aware attention module that helps the model decide which of a potentially infinite set of transactions to focus on, and significantly the most recent and in terms of time critical, which has implications for successfully identifying coordinated fraud. Moreover, TMR-GGNN has been developed as a guided contrastive decoder, thereby improving the quality of latent representations, which are easier to discriminate even in extreme class imbalance scenarios. Models such as SDT [12] achieve better accuracy but less recall, which means they have more frequent misses on fraud, whereas TMR-GGNN is designed to improve the trade-off, where Focal Loss is combined with InfoNCE to have two drivers of contrastive learning. While all these factors support the accuracy and efficacy of the proposed model, they also enhance its reliability, robustness, generalizability, and coverage of fraud, making it a more effective and trustworthy option for real-world fraud-detection systems.

## V. Conclusion

In this research, the proposed TMR-GGNN, which is a graph-based fraud-detection framework, effectively addresses key challenges such as complex entity interactions, temporal dynamics, and extreme class imbalance. Specifically, by constructing a heterogeneous transaction graph and integrating a time-aware relational attention mechanism with a guided contrastive decoder, the model captures both structural and temporal dependencies which are essential for fraud identification. Subsequently, the composite loss function is utilized to enhance the ability to distinguish minority class instances, resulting in superior performance across all key evaluation metrics compared with state-of-the-art models. The proposed TMR-GGNN attained greater results with 0.999 accuracy, 0.975 precision, 0.955 recall, 0.945 F1-score and 0.996 of AUC-ROC respectively. Hence, this architecture not only improves detection accuracy, but also ensures better fraud coverage and robustness. In future work, the proposed TMR-GGNN model will aim to extend the model to support real-time graph updates and incorporate explainability modules to improve transparency in high-stakes financial decision making.